\documentclass[11pt]{article}
\usepackage[T1]{fontenc}
\usepackage{geometry}
\usepackage{graphicx}
\usepackage{float}
\usepackage{capt-of}
\usepackage{csquotes}
\usepackage[
    style=apa,
    backend=biber,
    maxcitenames=2,
    uniquelist=false,
]{biblatex}
\usepackage{authblk}
\usepackage{amsmath}
\usepackage{xcolor}
\usepackage[colorlinks=true, linkcolor=blue, citecolor=blue, urlcolor=blue]{hyperref}
\usepackage[all]{hypcap}
\usepackage{abstract}

\newcommand{\beginsupplement}{%
    \setcounter{table}{0}
    \renewcommand{\thetable}{S\arabic{table}}%
    \setcounter{figure}{0}
    \renewcommand{\thefigure}{S\arabic{figure}}%
}

\title{Human-level 3D shape perception emerges from multi-view learning}
\vspace{-1em} 

\author[]{Tyler Bonnen\thanks{Corresponding author: bonnen@berkeley.edu }}
\author[]{Jitendra Malik}
\author[]{Angjoo Kanazawa}
\affil[]{University of California, Berkeley}
\begin{document}
\vspace{-1em}
\date{}
\maketitle
\vspace{-2em}
\begin{abstract}
\noindent Humans can infer the three-dimensional structure of objects from two-dimensional visual inputs. Modeling this ability has been a longstanding goal for the science and engineering of visual intelligence, yet decades of computational methods have fallen short of human performance. Here we develop a modeling framework that predicts human 3D shape inferences for arbitrary objects, directly from experimental stimuli. We achieve this with a novel class of neural networks trained using a visual-spatial objective over naturalistic sensory data; given a set of images taken from different locations within a natural scene, these models learn to predict spatial information related to these images, such as camera location and visual depth, without relying on any object-related inductive biases. Notably, these visual-spatial signals are analogous to sensory cues readily available to humans. We design a zero-shot evaluation approach to determine the performance of these `multi-view' models on a well established 3D perception task, then compare model and human behavior. Our modeling framework is the first to match human accuracy on 3D shape inferences, even without task-specific training or fine-tuning. Remarkably, independent readouts of model responses predict fine-grained measures of human behavior, including error patterns and reaction times, revealing a natural correspondence between model dynamics and human perception. Taken together, our findings indicate that human-level 3D perception can emerge from a simple, scalable learning objective over naturalistic visual-spatial data. Code, images, and human data needed to reproduce all analyses can be found at \href{https://tzler.github.io/human_multiview/}{https://tzler.github.io/human\_multiview/}

\end{abstract}

\section*{Introduction}
\vspace{.5em}
\noindent 
How do humans perceive the three-dimensional structure of objects? Cognitive scientists have assembled a rich body of empirical evidence to understand this visual ability. 3D shape inferences appear to rely on multiple visual processes that emerge early in infancy, yet still require years to fully develop (\cite{todd2004visual, yonas1987four, van2012keep}). This developmental process has been well characterized (\cite{piaget1948representation, gibson1969principles}), with recent work revealing the statistical structure of infants' natural experience: children have dense, visually diverse experiences with a relatively small number of objects and environments (\cite{smith2018developing, long2024babyview}), alongside rich multi-modal signals including depth from stereopsis and self-motion from the vestibular system (\cite{angelaki2008vestibular, campos2000travel}). While the sensory data and behavioral milestones typical for developing children are well characterized, the learning principles underlying 3D perception are not. One prominent hypothesis is that object perception emerges from general purpose learning mechanisms over these multi-sensory data distributions (\cite{von1867handbuch, kersten2004object}) while others have argued that object-related inductive biases are necessary for learning (\cite{spelke1990principles}). A longstanding goal within the cognitive sciences has been to design models that embody such hypotheses and evaluate them alongside human behavior (\cite{cao2024explanatory}). However, existing models have not achieved human-level 3D perception (\cite{bowers2023deep, bonnen2024evaluating}), limiting our ability to formally evaluate these cognitive theories.

\vspace{.5em}
\noindent
Here we develop a modeling framework that, for the first time, matches human performance on 3D perception tasks for arbitrary objects. We begin with a novel class of neural networks trained using visual and spatial information analogous to natural visual experience, including images, depth, and self-motion cues. These models learn via a conceptually simple and scalable learning objective: given a set of images captured from different locations within a scene, predict spatial information associated with each image, such as their relative locations or visual depth. This approach formulates perception as a correspondence problem, in which inferring image content is entangled with inferring spatial information; this is a stark contrast to prevailing modeling approaches within the cognitive sciences, which formulate vision as a feature extraction problem, and leverage powerful models trained on images sampled from unrelated environments. Critically, these visual-spatial models can be trained without any object-related inductive biases in the model architecture or learning objective (\cite{wang2025vggt}). As such, this modeling strategy embodies longstanding theories within the cognitive sciences that visual perception can emerge from general-purpose learning mechanisms operating over natural sensory experience (\cite{smith2005development}). Here we determine whether this approach achieves human-level 3D shape inferences by adapting these models for experimental tasks, then evaluating whether model responses predict human behavior.

\vspace{.5em}
\noindent 
To evaluate the alignment between humans and these `multi-view' vision models we leverage a perceptual task that requires zero-shot inferences about 3D shape. In this task, a single object is presented from two different viewpoints alongside a second object (i.e., images A, A', B) and participants must identify the non-matching object (B) using only visual information available in these three images. We develop a model evaluation framework that enables us to estimate their zero-shot performance on this task; we do not train models using this experimental setup, nor do we utilize any experimental stimuli/data for fine-tuning or training linear decoders. Instead, we leverage a confidence-based readout that is built into the model's training objective. While prior work reveals a considerable gap between humans and existing vision models (\cite{bonnen2024evaluating, bowers2023deep}), we substantially outperform prior models and match human-level accuracy. We develop complementary zero-shot analyses and demonstrate that independent model readouts predict human-like error patterns and reaction times---measures that have been the cornerstone of cognitive theories for decades. These findings reveal an emergent correspondence between multi-view models and human perception, indicating that human-level 3D shape inferences can emerge from a simple multi-view learning objective applied to naturalistic visual data.

\section*{Methods}

\subsection*{Multi-view vision models}

We evaluate a series of multi-view vision transformers including DUST3R (\cite{wang2024dust3r}), MAST3R (\cite{leroy2024grounding}), Pi3 (\cite{wang2025pi}), and VGGT-1B (\cite{wang2025vggt}). These models have been trained on large-scale, multi-view, naturalistic scene data (e.g., \cite{ling2024dl3dv}); these models receive sets of images depicting the same scene from different viewpoints and must learn to predict related spatial information, including camera location, visual depth, and correspondence (Fig. \ref{train_and_test} left). The training signals used in this modeling approach are analogous to information that is available to humans through stereo vision, tactile feedback, and proprioception. Notably, VGGT uses a generic transformer architecture (\cite{dosovitskiy2020image}) with no hand-coded geometric priors. As such, any understanding of 3D structure emerges from learning the predictive relationship between images and these multi-modal cues. We evaluate these models without any fine-tuning or task-specific adaptation (e.g., no linear decoders). Images were preprocessed by converting to RGB format, resizing to 518 pixels (with height adjusted to the nearest multiple of 14), and applying bicubic interpolation.

\subsection*{Experimental task}

\begin{figure}[ht!]
\begin{center}
\includegraphics[width=.95\textwidth]{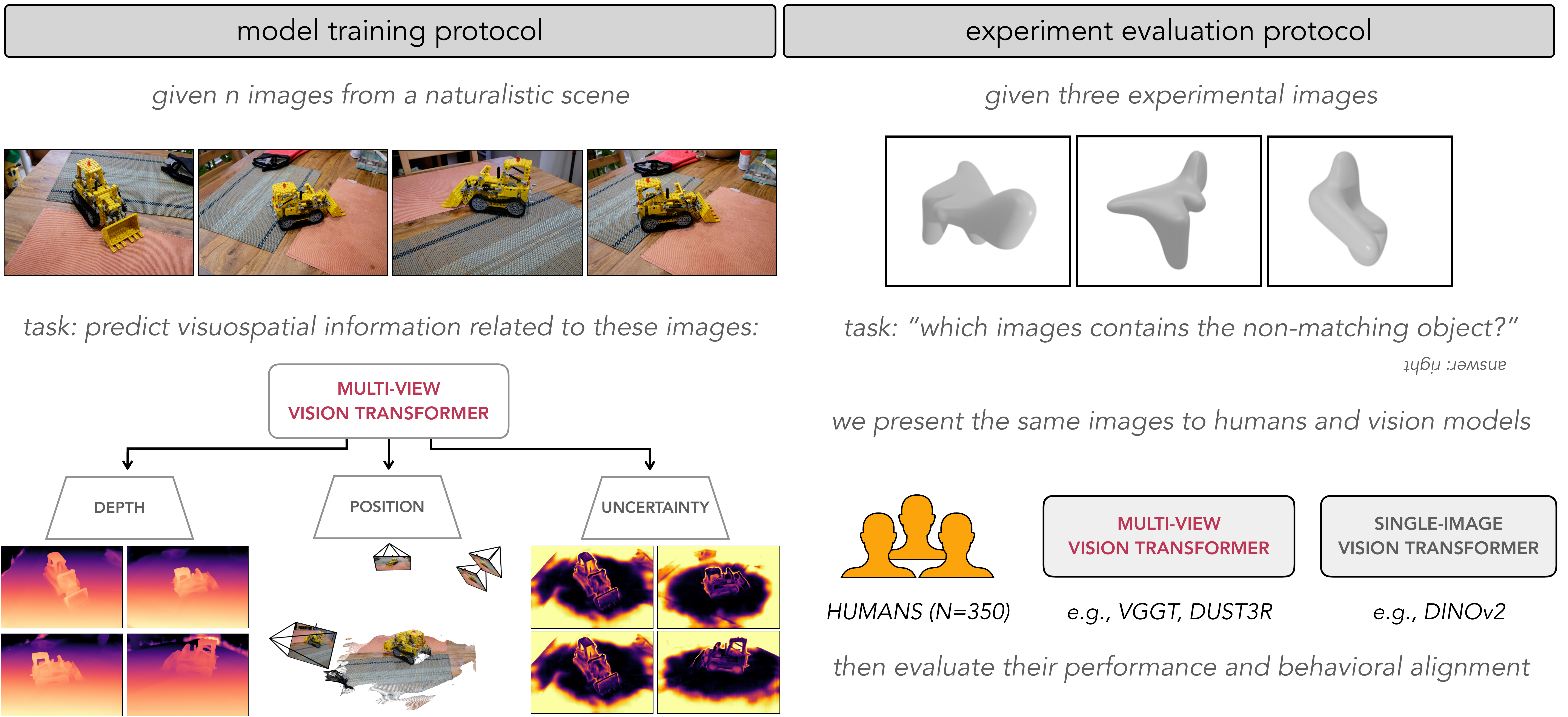}
\end{center}
\caption{\textbf{Schematic of multi-view model training approach and 3D perceptual testing protocol.} We evaluate a novel class of multi-view transformers (VGGT-1B, \cite{wang2025vggt}), which is trained on large-scale, multi-view, naturalistic scene data. During training, VGGT receives sets of images depicting the same scene from different viewpoints (top left) and must learn to predict the relative depth, camera position, and aleatoric uncertainty associated with these images (bottom left). These multi-modal signals are analogous to information that is available to humans through stereo vision and proprioception. Notably, VGGT uses a general transformer architecture with no hand-coded geometric priors: any understanding of 3D structure emerges from learning the predictive relationship between images and these multi-modal cues. To evaluate these multi-view transformers alongside human observers on a 3D perception task, we use a standard experimental design from the cognitive sciences: a concurrent visual discrimination (`oddity') task. This design requires zero-shot visual inference about object shape: given two images of an object from different viewpoints (A and A$'$), and another image of a different object (B), the task is to determine which image contains the non-matching object (e.g., right). We evaluate humans and models on diverse object types, including real-world objects (e.g., chairs, tables) as well as procedurally generated abstract shapes (i.e., `nonsense' objects).}
\label{train_and_test}
\end{figure}

Our experimental design requires zero-shot visual inference about object shape: given two images of an object from different viewpoints (A and A$'$), and another image of a different object (B), the task is to determine which image contains the non-matching object (see example trial in Fig. \ref{train_and_test} right). This design enables us to parametrically vary trial difficulty by changing the relative similarity between different objects (i.e., between object A and object B), and the viewpoint variation between images of the same object (i.e., between an image of object A and the same object from a different viewpoint, A$'$). Moreover, this task imposes minimal verbal demands, enabling us to focus on perceptual processes rather than language or semantic knowledge. We leverage experimental stimuli and behavioral data from a large-scale benchmark that has previously demonstrated a gap between humans and computer vision models (\cite{bonnen2024evaluating}). Images in this benchmark contain diverse object types and perceptual difficulties, including real-world objects (e.g., chairs, tables) as well as procedurally generated abstract shapes (i.e., `nonsense' objects without semantic attributes). 

\subsection*{Human behavioral data} 

We analyze human behavioral data from more than 300 human participants, collecting 25K trials of behavioral data online. Experimental data were collected online using Prolific. Participants were each paid \$15/hr for participating, and were free to terminate the experiment at any time. Experiments consisted of an initial set of instructions, 6 practice trials with feedback, and 150 main trials with no feedback. The 150 main trials were constructed such that no objects were repeated across trials, to avoid learning effects, and the ordering of the correct choice was randomized, to control for ordering effects. Given the performance we estimate for each image triplet, we normalize human accuracy to lie between zero (chance) and 1 (ceiling). This enables us to compare odd-one-out tasks and match-to-sample tasks in the same metric space. We collect eyetracking data on a subset of images and outline the data collection and analysis procedures for these data in the appendix.

\begin{figure}[ht!]
\begin{center}
\includegraphics[width=.85\textwidth]{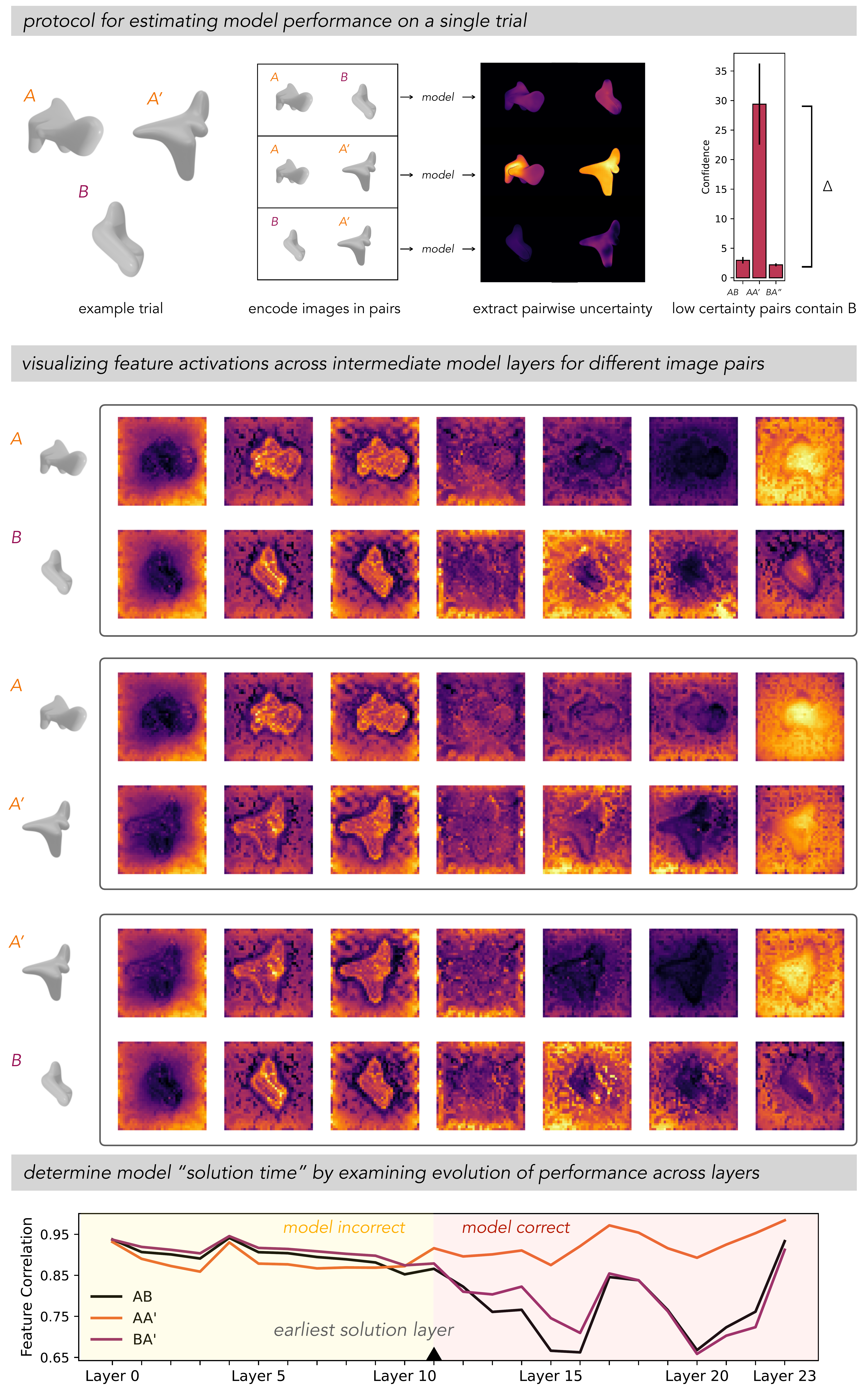}
\end{center}
\vspace{-5mm}
\caption{\textbf{Evaluation approach used to estimate model performance on each trial. } We develop a series of zero-shot evaluation metrics to determine the behavior of multi-view models on human 3D perceptual tasks. To estimate accuracy we leverage the model's internal estimate of aleatoric uncertainty (top): we encode all pairwise combinations of images on a trial, extract model uncertainty estimates for each pair, average, then select the oddity to be the object with the lowest paired confidence scores. We compare this model-selected non-match with the ground truth, resulting in a single binary (correct/incorrect) outcome. Next, we compute the margin between matching/non-matching objects to determine the model's confidence for this decision ($\Delta$ top right). We visualize the norm of model responses across several layers, going from early to final layers (left to right). Note that the activation patterns for matching (AA$'$) objects is visibly distinct from those of the non-matching (AB/BA$'$) objects in the final layers. When conducting a similarity analysis across all layers (bottom), for each encoded pair of images, we find that the features of matching objects (AA$'$) become more correlated (orange, bottom), while non-matching objects (AB/BA$'$) become less correlated (black/purple, bottom). We describe the earliest layer where the non-matching object can be identified as the model `solution layer'.    
}
\label{visualize_evaluations}
\end{figure}

\subsection*{A zero-shot metric to evaluate model performance}

To determine the behavior of multi-view models on human psychological tasks, we develop a series of independent, zero-shot evaluation metrics. First, in order to compare to human perceptual abilities, we estimate accuracy on this 3D perceptual task by leveraging the model's internal estimate of aleatoric uncertainty. During training, these models predicts per-pixel depth estimates along with uncertainty (precision) values using a loss function that weights prediction errors by confidence. The implementation varies by model, but for VGGT:

\begin{equation}
\mathcal{L}_{\text{depth}} = \sum_{i=1}^{N} \left\| \Sigma^D_i \odot (\hat{D}_i - D_i) \right\| + \left\| \Sigma^D_i \odot (\nabla\hat{D}_i - \nabla D_i) \right\| - \alpha \log \Sigma^D_i
\end{equation}

\noindent where $\Sigma^D_i$ is the predicted precision (inverse variance), $\hat{D}_i$ is predicted depth, and the gradient term enforces spatial smoothness. High precision regions correspond to reliable geometric correspondences, while low precision indicates ambiguity. We use these per-pixel uncertainty values in our analyses, as the uncertainty metric reflects the model's estimate of the geometric correspondence between image pairs. That is, when a given pixel belongs to the same location across two images, model confidence should be high, while pixels with no common origin across images should have low confidence. The model should have high `confidence' scores for image pairs with high correspondence (AA') and low confidence scores for images containing different objects (AB), even though it was never trained for this task. For all images within a given trial, we encode all pairwise combination of images, extract model uncertainty estimates for each pair, and select the non-matching object to be the image with the lowest confidence pairs. The simple assumption underlying this approach is that the non-matching object should yield lower geometric correspondence (and thus lower confidence) when paired with either of the matching objects. We then compare the model-selected non-match with the ground truth, resulting in a single binary (correct/incorrect) outcome for each trial. Refer to Fig. \ref{visualize_evaluations} (top) for a visualization of the protocol for a single example trial. Given our initial models evaluations (Fig. \ref{allmodels}) we restrict the subsequent analysis to VGGT. 

\subsection*{Estimating model behavioral confidence}

Next we develop a zero-shot evaluation metric to determine whether model confidence predicts human behavior beyond a coarse accuracy metric. As in the prior performance-based analysis, for images within a given trial we first encode all pairwise image combinations and extract model uncertainty estimates for each pair. Whereas the performance metric selects the non-matching object to be the image with the lowest confidence pairs, here we compute the margin between the matching and non-matching objects (visualized in Fig. \ref{readouts} top right: $\Delta$). To estimate this margin we average the confidence scores for the model-selected non-matching pairs (i.e., model-predicted AB and BA$'$) and subtract them from the confidence scores for the matching pair (AA$'$). This yields a single continuous value for each trial that corresponds to the margin separating the matching and non-matching objects that is agnostic to the correct/incorrect choice behavior.

\subsection*{Estimating model `solution time'}

Finally, we develop an independent measure to determine whether VGGT's computational dynamics mirror human perceptual processing. On each trial, we determined at which layer of the network the task could be reliably performed: the earliest layer where the model makes a correct oddity prediction that remains stable through all subsequent layers. This `model solution time' metric provides insight into when task-relevant visual representations emerge during VGGT's forward pass. For each trial, we extracted patch token representations from all 24 transformer layers in VGGT's aggregator (following the frozen DINOv2-Large encoder). 
We compute pairwise image similarities using three metrics at each layer: mean patch-to-patch cosine similarity, maximum patch-to-patch cosine similarity, and global pooling similarity (cosine similarity of spatially-averaged patch tokens). The predicted oddity was the image with minimum average similarity to all other images in the trial. 

\subsection*{Evaluating a baseline vision model}

To establish a baseline for the performance of vision models on these 3D perception tasks, we evaluate a state-of-the-art large image model (DINOv2-L; \cite{oquab2023dinov2}) which is also used as the encoder for VGGT. This provides a clear comparison between the features used as inputs to VGGT and the features of the multi-layer transformer that is learned from visual-spatial data. We estimate the non-matching object in each trial via a cosine distance metric: given model responses to images A, A', and B, we compute the pairwise similarity between items, then determine the `oddity' to be the item with the lowest off-diagonal similarity to the other images.

\subsection*{Qualitative visualization of model attention}

We provide qualitative visualizations of the information present in intermediate model layers by examining cross-image attention.  We visualize trials here by manually selected keypoints in the reference image, A (Fig. \ref{seeattention} far left), and given the target images for both A$'$ (Fig. \ref{seeattention} left, above) and B (Fig. \ref{seeattention}, bottom) we visualize the corresponding attention maps for each query point from A (Fig. \ref{seeattention}, right). For a given trial (e.g., Fig. \ref{seeattention}), we encoded image pairs A, A$'$ (Fig. \ref{seeattention}, top) and A, B (Fig. \ref{seeattention}, bottom) separately, then extracted the cross-image block of the attention matrix: for each selected source location on image A (visualized as different colored circles in Fig. \ref{seeattention}, left), we identified the corresponding patch token and retrieved its attention distribution over all patch tokens in the target image. We averaged attention weights across all 16 heads within a layer, upsampled the resulting 37$\times$37 attention map to image resolution via bilinear interpolation, and applied Gaussian smoothing. We masked the background using a luminance threshold on the target image. Attention maps for each source point were normalized to a common color scale across target images (A$'$ and B), enabling direct comparison of how the model distributes attention to matching versus non-matching objects from the same source location.

\section*{Results}

\subsection*{Multi-view vision models achieve human-level 3D perceptual accuracy}

We compared the performance of multi-view vision models, single-view image models, and humans on a concurrent visual discrimination task requiring 3D shape perception. For each trial, we determined whether the model correctly identified the non-matching object and compared this to human accuracy on the same trials. We analyzed behavioral data from more than 300 human participants across thousands of trials containing diverse object types, including familiar real-world objects (e.g., chairs, tables) and procedurally generated abstract shapes without semantic attributes. We normalized all performance metrics relative to chance performance for each dataset before statistical comparison such that accuracy ranged from zero (chance) to 1 (ceiling). 

\begin{figure}[ht!]
\begin{center}
\includegraphics[width=1\textwidth]{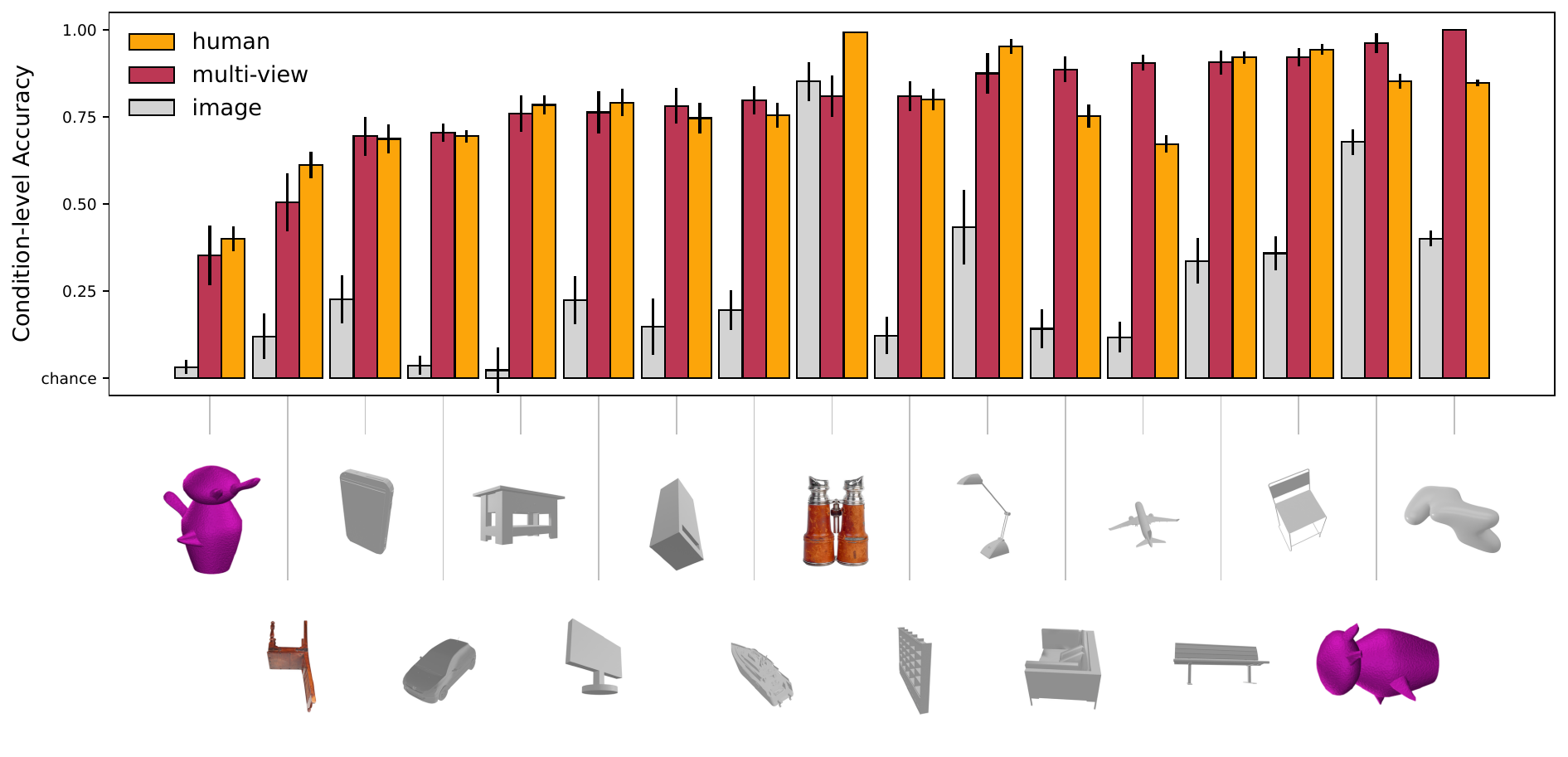}
\end{center}
\vspace{-5mm}
\caption{\textbf{Condition-level performance of humans and vision models.} We visualize the performance of humans (orange), multi-view vision models (red), and image models (grey) for each individual condition on this benchmark. These conditions contain both abstract (e.g., 'greebles (high similarity)'; far left) and real-world objects (e.g., 'familiar objects (low similarity)' far left) rendered in color and greyscale. Across these abstract and real-world conditions there is a close correspondence between multi-view models and human observers, while both substantially outperform image models. We also visualize a single image from a trial in each of these conditions which are, from left to right: 'greebles (high similarity)', 'familiar objects (high similar)', 'telephone', 'car', 'table', 'display', 'loudspeaker', 'watercraft', 'familiar objects (low similarity)', 'cabinet', 'lamp', 'sofa', 'airplane', 'bench', 'chair', 'greebles (low similarity)', and 'abstract objects'. }
\label{seeconditions}
\end{figure}

\vspace{.5em}
\noindent VGGT achieved an average normalized accuracy of 83.0\% ($\pm$ 3.7\% SEM), matching human performance (78.9\% $\pm$ 3.0\% SEM). A paired $t$-test on normalized performance across 21 dataset$\times$condition combinations revealed no significant difference between human and VGGT accuracy ($t$(20) = $-$1.67, $p$ = 0.110). This represents a dramatic improvement over single-view image models: DINOv2-Large, which serves as VGGT's visual encoder, achieved only 28.5\% ($\pm$ 4.9\% SEM) normalized accuracy---substantially below both human and VGGT performance (VGGT vs. DINOv2: $t$(20) = 12.28, $p$ $<$ 0.001, Cohen's $d$ = 2.68). This human-model correspondence is evident across stimulus conditions of all types, including `abstract' and 'semantic,' real-world and procedurally generated objects (Fig. \ref{seeconditions}). Critically, VGGT's human-level performance does not depend on task-specific training or fine-tuning; the model used only its pre-trained representations learned from multi-view prediction. These findings demonstrate that multi-view learning is sufficient to achieve human-level 3D object perception. 

\vspace{1em}
\begin{figure}[ht!]
\begin{center}
\includegraphics[width=.95\textwidth]{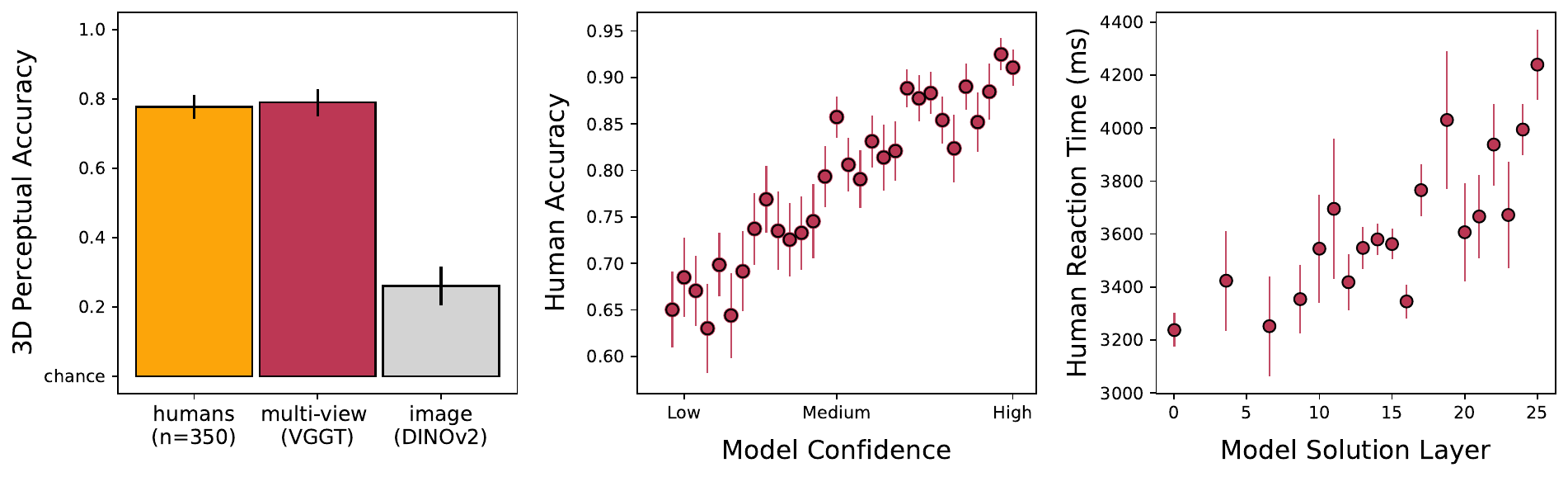}
\end{center}
\vspace{-5mm}
\caption{\textbf{Multi-view models match human 3D perception accuracy, error patterns, and reaction times.} 
When comparing normalized accuracy of humans and models across all conditions of this 3D perception benchmark (left), VGGT matches human performance, while both humans and multi-view models significantly outperform standard vision models like DINOv2. Critically, VGGT's human-level performance does not depend on task-specific training or fine-tuning; the model used only its pre-trained representations learned from multi-view prediction. These findings demonstrate that multi-view learning is sufficient to achieve human-level 3D object perception. Beyond average accuracy, we find that model confidence (i.e., the margin between matching and non-matching objects in each trial) is significantly correlated with human choice behaviors. This indicates that the aleatoric uncertainty used during training provides a natural analogue for human perceptual judgments. Finally, we observe a clear correspondence between model solution layer and human reaction time; as the number of layers required to solve this perceptual task increases, so too does human reaction time needed for correct responses. These data reveal an emergent correspondence between model dynamics and human perception.}
\label{mainresults}
\end{figure}

\subsection*{Model confidence predicts human perceptual difficulty}

Beyond overall accuracy, we investigated whether VGGT's internal representations capture the graded difficulty humans experience across trials. For each trial, we computed a confidence margin by comparing VGGT's uncertainty estimates for matching versus non-matching object pairs---a measure derived from the model's depth prediction precision that was never optimized for predicting human behavior. We then binned trials by model confidence (30 quantile bins) and examined whether human accuracy varied systematically across these bins.

\vspace{.5em}
\noindent 
Human accuracy was strongly predicted by model confidence (Pearson $r$ = 0.830, $p$ $<$ 0.001; Spearman $\rho$ = 0.932, $p$ $<$ 0.001). We observe a clear correspondence between model confidence and human accuracy; as VGGT's confidence increases, so too does human accuracy (OLS regression $\beta$ = 0.01, $F$(1, 28) = 61.92, $p$ = 1.43e-08). Trials where VGGT had low confidence (lowest confidence bin) were substantially more difficult for humans, with mean accuracy of 63.0\%. Conversely, trials where VGGT had high confidence (highest confidence bin) corresponded to trials where humans performed near ceiling (92.5\% accurate). This monotonic relationship between model confidence and human accuracy, spanning a range of 29.5\% percentage points ($R^2$ = 0.689), reveals a striking correspondence in how the model and humans evaluate task difficulty. The fact that this internal model variable independently predicts human performance patterns suggests an emergent alignment between VGGT and human 3D perception.

\subsection*{Model solution dynamics predict human reaction times}

Finally, we examined whether the temporal dynamics of VGGT's computations correspond to the time humans require to make perceptual decisions. For each trial, we determined the `solution layer'---the earliest layer in VGGT's 24-layer transformer where the correct oddity decision emerged and remained stable through all subsequent layers. This solution layer metric provides insight into the depth of processing required for each trial: trials solvable in early layers require minimal computation, while difficult trials require processing through deeper layers. We hypothesized that trials requiring deeper processing in the model (later solution layers) would require longer reaction times in humans.

\vspace{.5em}
\noindent 
We find a clear correspondence between model solution layer and human reaction time (Pearson $r$ = 0.796, $p$ $<$ 0.001; Spearman $\rho$ = 0.804, $p$ $<$ 0.001); as the depth of processing in VGGT increases, so too does human RT (OLS regression $\beta$ = 30.91 ms/layer, $F$(1, 17) = 29.38, $p$ = 4.59e-05). Trials that VGGT solved in early layers corresponded to fast human responses (minimum bin mean RT = 3238 ms), while trials requiring processing through later layers were associated with substantially slower human responses (maximum bin mean RT = 4240 ms), spanning a range of 1002 ms ($R^2$ = 0.633). Critically, this relationship remained significant even when controlling for model confidence (partial correlation: $r$ = 0.159, $p$ = 2.78e-10), indicating that the solution layer captures aspects of processing dynamics independent of overall task difficulty. These findings reveal a correspondence not just in what decisions humans and VGGT make (accuracy) or how difficult they find trials (confidence), but in the temporal dynamics of how these decisions unfold. The depth of processing required within VGGT's feedforward architecture naturally maps onto the time humans require for perceptual inference.

\begin{figure}[ht!]
\begin{center}
\includegraphics[width=1\textwidth]{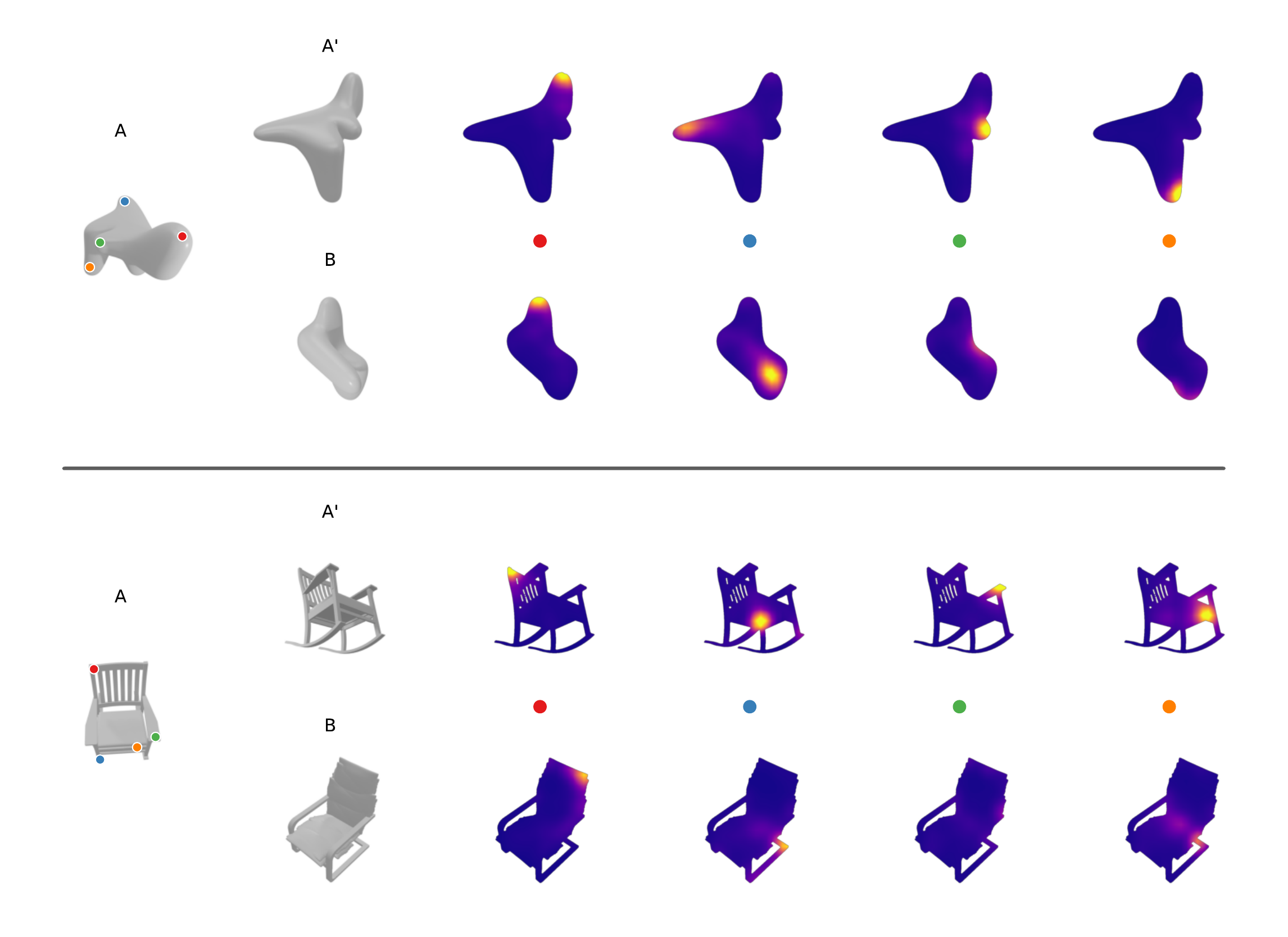}
\end{center}
\vspace{-5mm}
\caption{\textbf{Investigating between-image activations reveals location-based object correspondence.} How does this multi-view model represent the similarity or difference between objects? To address this question we provide a qualitative visualization of the information present in intermediate model layers. For an example trial (e.g., abstract image in the top half; chair in the bottom half), we encoded image pairs A, A$'$ and A, B separately, then extract the cross-image block of the attention matrix. Here we manually select keypoints in the reference image, A (different color dots on A, far left), and identify the corresponding patch token in the target images  A$'$ and B. We then retrieve the attention distribution over all the target patch tokens. In intermediate model layers (e.g., here we visual attention from layer 15) we find that different query locations from the reference image A elicit distinct attention patterns across points and across target images A$'$ and B. Concretely, it appears that each query location on A elicits a pattern of attention in A$'$ that correspondence to the same location on the object, albeit a different location in xyz coordinates. This qualitative analysis indicates that the model can represent the object-object similarity via the correspondence between spatial locations on each object. }
\label{seeattention}
\end{figure}

\subsection*{Qualitative analysis of model attention reveals correspondence-based representations}

\begin{figure}[ht!]
\begin{center}
\includegraphics[width=1\textwidth]{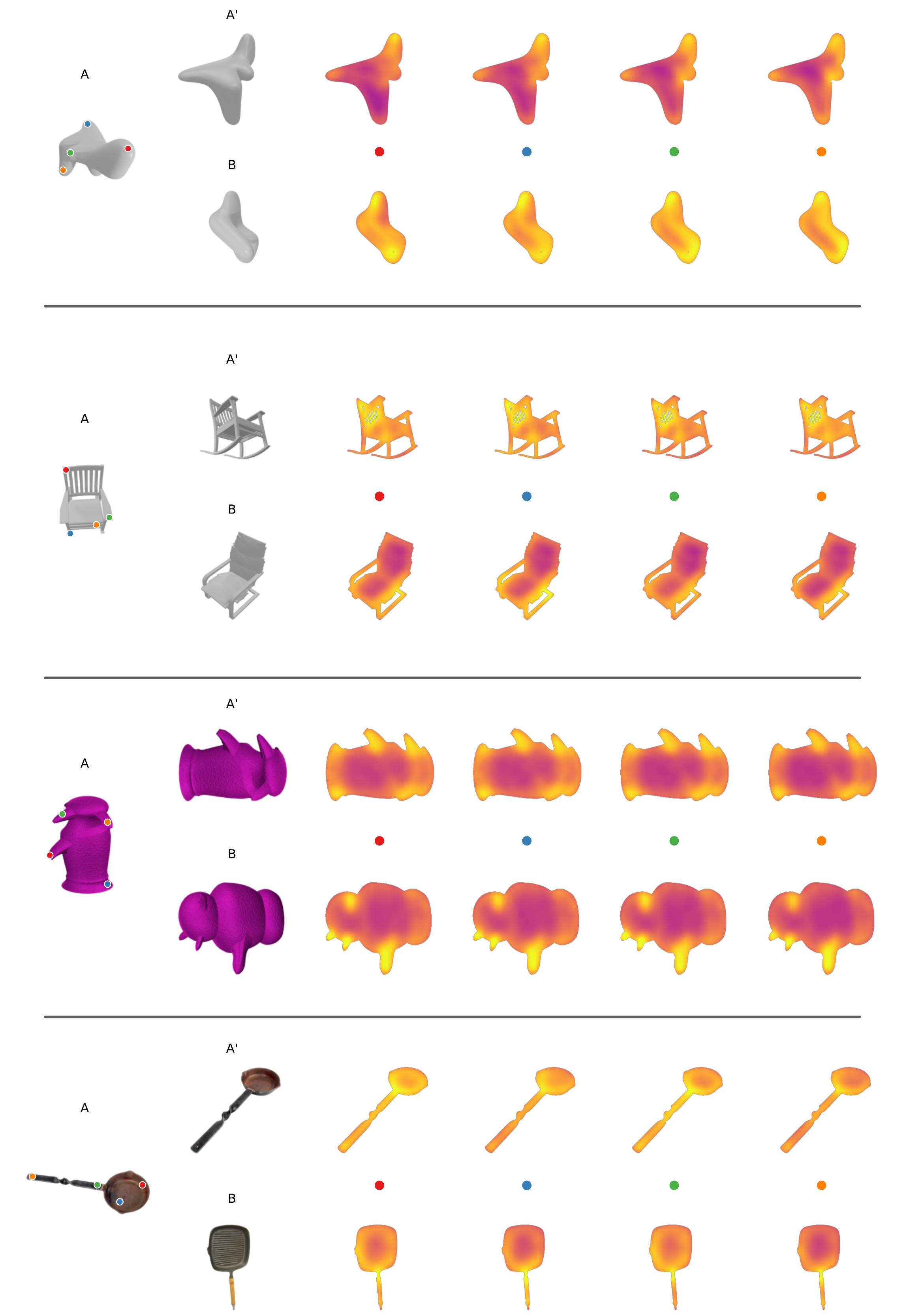}
\end{center}
\vspace{-5mm}
\caption{\textbf{Pattern of global attention across different objects in layer zero.}}
\label{seeattention_early}
\end{figure}

How does this multi-view model represent the similarity or difference between objects in this 3D shape perception task? To address this question we provide a qualitative visualization of the information present in intermediate model layers. For a given trial, we encoded image pairs A, A$'$ and A, B separately, then extract the cross-image block of the attention matrix. For each selected source location on image A, we identified the corresponding patch token and retrieved its attention distribution over all patch tokens in the target image. Here we manually select keypoints in the reference image, A, and given the target images (for A$'$ and B, separately) we visualize the corresponding attention maps for each query point from A. In early layers (Fig. \ref{seeattention_early}) there is little differentiation between locations (Fig. \ref{seeattention_early} across columns) or between objects (Fig. \ref{seeattention_early} across rows). However, in intermediate model layers (e.g., layer 15, Fig. \ref{seeattention}) we find that different query locations from the reference image A (visualized as different colored circles in Fig. \ref{seeattention}, left) elicit distinct attention patterns (Fig. \ref{seeattention}, right) and across target images A$'$ and B (Fig. \ref{seeattention}, top row vs bottom for each triplet). It appears that each query location on A elicits a pattern of attention in A$'$ that corresponds to the same location on the object, albeit a different location in xyz coordinates (e.g., for both the abstract and semantic object triplets in Fig. \ref{seeattention}, attention maps on the target location on A project to the same object part in A$'$). This qualitative analysis indicates that the model can represent the object-object similarity via the correspondence between spatial locations on each object. 

\vspace{.5em}

\section*{Discussion}

\noindent
We have introduced the first modeling framework that approximates human performance on 3D visual object perception tasks. Leveraging a novel class of `multi-view' models from the computer sciences, we developed an evaluation framework that enables us to compare model and human behavior zero-shot. When presented with the same images as experimental participants, the model achieves human-level accuracy and predicts fine-grained patterns of behavior, including error patterns and reaction time. Critically, we predict these behavioral measures without any task-specific training or fine-tuning. This approach builds off of decades of modeling efforts across multiple disciplines (\cite{fukushima1980neocognitron, olshausen1996emergence, yamins2016using}). Notably, there has been a recent push to model 3D perceptual abilities using a diversity of strategies (\cite{pandey2025computational, khazoum2025deep, o2025approximating, leebiologically}). Nonetheless, when these modeling approaches were evaluated alongside human behavior, they were only capable of matching human accuracy for objects that they were trained on---i.e., they were not capable of the generalization abilities that are emblematic of human perception. Our work bridges the gap between humans and computational models by formulating vision as a sequential visual-spatial inference, which is learned by training on large-scale naturalistic data. In doing so, our work realizes a longstanding goal of the cognitive sciences: to construct computational models of human 3D perceptual inferences that are capable of formalizing and evaluating theories of human perception. 

\vspace{.5em}
\noindent
Multi-view models offer several advantages over the prevailing vision encoders that are used throughout the cognitive and neurosciences of vision. These multi-view models are designed to process visual sequences in a manner that accommodates the sequential, multi-modal data that more closely approximate the sensory distributions animals experience in natural environments. This provides novel opportunities to experiment with different biologically plausible training recipes and evaluation frameworks. For example, these models can incorporate spatial information (e.g., camera position, depth) that is readily available to humans through proprioception, stereo vision, and active exploration. Given the emergent alignment between this modeling framework and human perception, we can begin to disentangle the relative contributions that different sensory modalities (e.g., visual depth vs. self motion) might have for human perceptual development. Beyond experimenting with different training strategies, the multi-view nature of these vision models introduces a relational component to visual processing that is absent in image models, providing novel opportunities to formalize the relational encoding dynamics that are inherent in visual processes. For example, prior modeling approaches have---in practice---assumed that a model's 3D shape understanding must be evaluated by extracting features from each image independently. In contrast, we have formulated 3D shape understanding (both in terms of model training, as well as our evaluation framework) as a process of integrating information across multiple images. In our evaluations, for example, we used a pair-wise encoding strategy, but there is a rich space of design choices here for experimentalists to explore. 

\vspace{.5em}
\noindent
While these multi-view models match human behavioral patterns, future work is needed to ensure that they reflect the neural structures and algorithms that support human perceptual abilities. For example, it is well established that human reaction times on 3D shape inferences emerge from complex dynamics that include sequential eye movements and active information sampling (\cite{ullman1979interpretation}). In contrast, our solution layer metric reflects depth of feedforward processing in a single pass through the network---entirely lacking the foveal constraints that are emblematic of human vision. As such, the temporal correspondence we observe may reflect shared computational demands (e.g., task difficulty) rather than shared algorithmic solution. Architecturally, this modeling approach already reflects some of the large-scale neuro-anatomical motifs that are thought to support visual perception: a feedforward encoder passing information to a downstream structure that is able to sequentially integrate these outputs (\cite{bonnen2025medial, bonnen2021ventral}). While these feedforward dynamics can be thought of as an over-parametrized recurrent neural network (\cite{jacobs2025block}), future work is needed to ensure that these operations reflect what is known about neural structures and dynamics underlying human perception. Finally, we note that while VGGT's inputs are analogous to human sensory data, the correspondence is imperfect; the positional information that these models process is in a global (not egocentric self-motion) coordinate frame, and this model receives depth supervision that is not faithful to stereoscopic cues. These gaps between model and human processing present exciting opportunities for developing next-generation models with stronger biological constraints.

\vspace{.5em}
\noindent
The science and engineering of intelligence has grappled with a common question for decades: do sophisticated abilities emerge from domain-general learning strategies, or do they require built-in, domain-specific knowledge? In the cognitive science of vision, `empiricist' accounts have argued that perception emerges from general-purpose learning algorithms over structured sensory experience (\cite{fiser2002statistical}), while `nativist' accounts proposed that object-specific inductive biases are necessary for learning (\cite{spelke1990principles}). The computer sciences echoed this same divide, with computer vision developing learning-based strategies to represent the visual world (\cite{lecun2002gradient}), while computer graphics designing models with built-in geometric knowledge (\cite{hartley2003multiple}). The recent success of `multi-view' modeling approaches, which learn about the 3D world without any hard-coded geometric knowledge (\cite{wang2025vggt}), reflects a broader realization in the computer sciences: scalable, self-supervised learning strategies outperform models with domain-specific inductive biases (\cite{sutton2019bitter}). Our results suggest that these engineering innovations have real implications for cognitive theory. Not only are these models competitive on computer science benchmarks, but they exhibit an emergent correspondence with human perception. These results corroborate `empiricist' theories of cognition that emphasize the data-dependent nature of human intelligence (\cite{mcclelland2010letting, elman1996rethinking}), while providing new methodological frameworks to formalize and evaluate our theories of human cognition. 

\section*{Acknowledgements}

This work is supported by the National Institute of Neurological Disorders and Stroke within the NIH (Award Number F99NS125816) and the UC Presidential Postdoctoral Fellowship Award. We thank Sophia Koepke, Stephanie Fu, Amil Dravid, Konpat Preechakul, Alyosha Efros, Mark Ho, Chris Iyer, and Andrew Lampinen,  for fruitful discussions and manuscript feedback.  

\section*{Code and data availability}

All code, images, and human behavioral data needed to reproduce all results can be found on our \href{https://tzler.github.io/human_multiview/}{project page}. Model analysis can be found on \href{https://github.com/tzler/human_multiview/}{github}. Experimental images and human behavioral data are on \href{https://huggingface.co/datasets/tzler/MOCHI}{huggingface}.

\printbibliography

@article{mcclelland2010letting,
  title={Letting structure emerge: connectionist and dynamical systems approaches to cognition},
  author={McClelland, James L and Botvinick, Matthew M and Noelle, David C and Plaut, David C and Rogers, Timothy T and Seidenberg, Mark S and Smith, Linda B},
  journal={Trends in Cognitive Sciences},
  volume={14},
  number={8},
  pages={348--356},
  year={2010},
  publisher={Elsevier}
}

@article{yamins2016using,
  title={Using goal-driven deep learning models to understand sensory cortex},
  author={Yamins, Daniel LK and DiCarlo, James J},
  journal={Nature Neuroscience},
  volume={19},
  number={3},
  pages={356--365},
  year={2016},
  publisher={Nature Publishing Group}
}

@article{pandey2025computational,
  title={Computational origins of shape perception},
  author={Pandey, Lalit and Wood, Samantha MW and Wood, Justin N},
  journal={PLOS Computational Biology},
  volume={21},
  number={12},
  pages={e1013674},
  year={2025},
  publisher={Public Library of Science San Francisco, CA USA}
}

@article{smith2005development,
  title={The development of embodied cognition: Six lessons from babies},
  author={Smith, Linda and Gasser, Michael},
  journal={Artificial life},
  volume={11},
  number={1-2},
  pages={13--29},
  year={2005},
  publisher={MIT Press}
}

@article{bonnen2021ventral,
  title={When the ventral visual stream is not enough: A deep learning account of medial temporal lobe involvement in perception},
  author={Bonnen, Tyler and Yamins, Daniel LK and Wagner, Anthony D},
  journal={Neuron},
  volume={109},
  number={17},
  pages={2755--2766},
  year={2021},
  publisher={Elsevier}
}

@article{bowers2023deep,
  title={Deep problems with neural network models of human vision},
  author={Bowers, Jeffrey S and Malhotra, Gaurav and Dujmovi{\'c}, Marin and Montero, Milton Llera and Tsvetkov, Christian and Biscione, Valerio and Puebla, Guillermo and Adolfi, Federico and Hummel, John E and Heaton, Rachel F and others},
  journal={Behavioral and Brain Sciences},
  volume={46},
  pages={e385},
  year={2023},
  publisher={Cambridge University Press}
}

@article{o2025approximating,
  title={Approximating Human-Level 3D Visual Inferences With Deep Neural Networks},
  author={O’Connell, Thomas P and Bonnen, Tyler and Friedman, Yoni and Tewari, Ayush and Sitzmann, Vincent and Tenenbaum, Joshua B and Kanwisher, Nancy},
  journal={Open Mind},
  volume={9},
  pages={305--324},
  year={2025},
  publisher={MIT Press 255 Main Street, 9th Floor, Cambridge, Massachusetts 02142, USA~…}
}

@article{bonnen2024evaluating,
  title={Evaluating multiview object consistency in humans and image models},
  author={Bonnen, Tyler and Fu, Stephanie and Bai, Yutong and O'Connell, Thomas and Friedman, Yoni and Kanwisher, Nancy and Tenenbaum, Joshua B and Efros, Alexei A},
  journal={Advances in Neural Information Processing Systems},
  volume={37},
  pages={43533--43548},
  year={2024}
}

@book{hartley2003multiple,
  title={Multiple view geometry in computer vision},
  author={Hartley, Richard and Zisserman, Andrew},
  year={2003},
  publisher={Cambridge university press}
}

@article{dosovitskiy2020image,
  title={An image is worth 16x16 words: Transformers for image recognition at scale},
  author={Dosovitskiy, Alexey},
  journal={arXiv preprint arXiv:2010.11929},
  year={2020}
}

@book{elman1996rethinking,
  title={Rethinking innateness: A connectionist perspective on development},
  author={Elman, Jeffrey L},
  volume={10},
  year={1996},
  publisher={MIT press}
}

@article{spelke1990principles,
  title={Principles of object perception},
  author={Spelke, Elizabeth S},
  journal={Cognitive science},
  volume={14},
  number={1},
  pages={29--56},
  year={1990},
  publisher={Elsevier}
}

@inproceedings{wang2025vggt,
  title={Vggt: Visual geometry grounded transformer},
  author={Wang, Jianyuan and Chen, Minghao and Karaev, Nikita and Vedaldi, Andrea and Rupprecht, Christian and Novotny, David},
  booktitle={Proceedings of the Computer Vision and Pattern Recognition Conference},
  pages={5294--5306},
  year={2025}
}

@inproceedings{wang2024dust3r,
  title={Dust3r: Geometric 3d vision made easy},
  author={Wang, Shuzhe and Leroy, Vincent and Cabon, Yohann and Chidlovskii, Boris and Revaud, Jerome},
  booktitle={Proceedings of the IEEE/CVF Conference on Computer Vision and Pattern Recognition},
  pages={20697--20709},
  year={2024}
}

@article{oquab2023dinov2,
  title={Dinov2: Learning robust visual features without supervision},
  author={Oquab, Maxime and Darcet, Timoth{\'e}e and Moutakanni, Th{\'e}o and Vo, Huy and Szafraniec, Marc and Khalidov, Vasil and Fernandez, Pierre and Haziza, Daniel and Massa, Francisco and El-Nouby, Alaaeldin and others},
  journal={arXiv preprint arXiv:2304.07193},
  year={2023}
}

@article{fukushima1980neocognitron,
  title={Neocognitron: A self-organizing neural network model for a mechanism of pattern recognition unaffected by shift in position},
  author={Fukushima, Kunihiko},
  journal={Biological cybernetics},
  volume={36},
  number={4},
  pages={193--202},
  year={1980},
  publisher={Springer}
}

@article{lecun2002gradient,
  title={Gradient-based learning applied to document recognition},
  author={LeCun, Yann and Bottou, L{\'e}on and Bengio, Yoshua and Haffner, Patrick},
  journal={Proceedings of the IEEE},
  volume={86},
  number={11},
  pages={2278--2324},
  year={2002},
  publisher={Ieee}
}

@article{piaget1948representation,
  title={La repr{\'e}sentation de l'espace chez l'enfant.},
  author={Piaget, Jean and Inhelder, B{\"a}rbel},
  year={1948},
  publisher={Presses universitaires de France}
}

@article{gibson1969principles,
  title={Principles of perceptual learning and development.},
  author={Gibson, Eleanor J},
  year={1969},
  publisher={Appleton-Century-Crofts}
}

@article{bonnen2025medial,
  title={Medial temporal cortex supports object perception by integrating over visuospatial sequences},
  author={Bonnen, Tyler and Wagner, Anthony D and Yamins, Daniel LK},
  journal={Cognition},
  volume={262},
  pages={106135},
  year={2025},
  publisher={Elsevier}
}

@article{sutton2019bitter,
  title={The bitter lesson},
  author={Sutton, Richard},
  journal={Incomplete Ideas (blog)},
  volume={13},
  number={1},
  pages={38},
  year={2019}
}

@article{ullman1979interpretation,
  title={The interpretation of structure from motion},
  author={Ullman, Shimon},
  journal={Proceedings of the Royal Society of London. Series B. Biological Sciences},
  volume={203},
  number={1153},
  pages={405--426},
  year={1979},
  publisher={The Royal Society London}
}

@book{von1867handbuch,
  title={Handbuch der physiologischen Optik},
  author={Von Helmholtz, Hermann},
  volume={9},
  year={1867},
  publisher={L. Voss}
}

@inproceedings{ling2024dl3dv,
  title={Dl3dv-10k: A large-scale scene dataset for deep learning-based 3d vision},
  author={Ling, Lu and Sheng, Yichen and Tu, Zhi and Zhao, Wentian and Xin, Cheng and Wan, Kun and Yu, Lantao and Guo, Qianyu and Yu, Zixun and Lu, Yawen and others},
  booktitle={Proceedings of the IEEE/CVF Conference on Computer Vision and Pattern Recognition},
  pages={22160--22169},
  year={2024}
}

@article{yonas1987four,
  title={Four-month-old infants' sensitivity to binocular and kinetic information for three-dimensional-object shape},
  author={Yonas, Albert and Arterberry, Martha E and Granrud, Carl E},
  journal={Child Development},
  pages={910--917},
  year={1987},
  publisher={JSTOR}
}

@article{van2012keep,
  title={Keep your eyes on development: the behavioral and neurophysiological development of visual mechanisms underlying form processing},
  author={Van Den Boomen, Carlijn and van der Smagt, Maarten J and Kemner, Chantal},
  journal={Frontiers in Psychiatry},
  volume={3},
  pages={16},
  year={2012},
  publisher={Frontiers Research Foundation}
}

@article{smith2018developing,
  title={The developing infant creates a curriculum for statistical learning},
  author={Smith, Linda B and Jayaraman, Swapnaa and Clerkin, Elizabeth and Yu, Chen},
  journal={Trends in cognitive sciences},
  volume={22},
  number={4},
  pages={325--336},
  year={2018},
  publisher={Elsevier}
}

@article{todd2004visual,
  title={The visual perception of 3D shape},
  author={Todd, James T},
  journal={Trends in cognitive sciences},
  volume={8},
  number={3},
  pages={115--121},
  year={2004},
  publisher={Elsevier}
}

@article{olshausen1996emergence,
  title={Emergence of simple-cell receptive field properties by learning a sparse code for natural images},
  author={Olshausen, Bruno A and Field, David J},
  journal={Nature},
  volume={381},
  number={6583},
  pages={607--609},
  year={1996},
  publisher={Nature Publishing Group UK London}
}

@article{khazoum2025deep,
  title={A Deep Learning Model of Mental Rotation Informed by Interactive VR Experiments},
  author={Khazoum, Raymond and Fernandes, Daniela and Krylov, Aleksandr and Li, Qin and Deny, Stephane},
  journal={arXiv preprint arXiv:2512.13517},
  year={2025}
}

@article{campos2000travel,
  title={Travel broadens the mind},
  author={Campos, Joseph J and Anderson, David I and Barbu-Roth, Marianne A and Hubbard, Edward M and Hertenstein, Matthew J and Witherington, David},
  journal={Infancy},
  volume={1},
  number={2},
  pages={149--219},
  year={2000},
  publisher={Taylor \& Francis}
}

@article{kersten2004object,
  title={Object perception as Bayesian inference},
  author={Kersten, Daniel and Mamassian, Pascal and Yuille, Alan},
  journal={Annu. Rev. Psychol.},
  volume={55},
  number={1},
  pages={271--304},
  year={2004},
  publisher={Annual Reviews}
}

@article{fiser2002statistical,
  title={Statistical learning of higher-order temporal structure from visual shape sequences.},
  author={Fiser, J{\'o}zsef and Aslin, Richard N},
  journal={Journal of Experimental Psychology: Learning, Memory, and Cognition},
  volume={28},
  number={3},
  pages={458},
  year={2002},
  publisher={American Psychological Association}
}

@inproceedings{leroy2024grounding,
  title={Grounding image matching in 3d with mast3r},
  author={Leroy, Vincent and Cabon, Yohann and Revaud, J{\'e}r{\^o}me},
  booktitle={European Conference on Computer Vision},
  pages={71--91},
  year={2024},
  organization={Springer}
}

@article{wang2025pi,
  title={$\pi^{3}$: Permutation-Equivariant Visual Geometry Learning},
  author={Wang, Yifan and Zhou, Jianjun and Zhu, Haoyi and Chang, Wenzheng and Zhou, Yang and Li, Zizun and Chen, Junyi and Pang, Jiangmiao and Shen, Chunhua and He, Tong},
  journal={arXiv preprint arXiv:2507.13347},
  year={2025}
}

@article{jacobs2025block,
  title={Block-Recurrent Dynamics in Vision Transformers},
  author={Jacobs, Mozes and Fel, Thomas and Hakim, Richard and Brondetta, Alessandra and Ba, Demba and Keller, T Andy},
  journal={arXiv preprint arXiv:2512.19941},
  year={2025}
}

@article{leebiologically,
  title={A biologically plausible route to learn 3D perception},
  author={Lee, Wanhee and Watrous, Jared and Chen, Honglin and Kotar, Klemen and Bonnen, Tyler and Yamins, Daniel LK}, 
  year={2024}
}

@article{cao2024explanatory,
  title={Explanatory models in neuroscience, Part 1: Taking mechanistic abstraction seriously},
  author={Cao, Rosa and Yamins, Daniel},
  journal={Cognitive Systems Research},
  volume={87},
  pages={101244},
  year={2024},
  publisher={Elsevier}
}

@article{angelaki2008vestibular,
  title={Vestibular system: the many facets of a multimodal sense},
  author={Angelaki, Dora E and Cullen, Kathleen E},
  journal={Annu. Rev. Neurosci.},
  volume={31},
  number={1},
  pages={125--150},
  year={2008},
  publisher={Annual Reviews}
}

@article{long2024babyview,
  title={The BabyView dataset: High-resolution egocentric videos of infants' and young children's everyday experiences},
  author={Long, Bria and Sparks, Robert Z and Xiang, Violet and Stojanov, Stefan and Yin, Zi and Keene, Grace E and Tan, Alvin WM and Feng, Steven Y and Zhuang, Chengxu and Marchman, Virginia A and others},
  journal={arXiv preprint arXiv:2406.10447},
  year={2024}
}

\newpage
\section{Supplementary Material}
\beginsupplement

\begin{figure}[ht!]
\begin{center}
\includegraphics[width=1\textwidth]{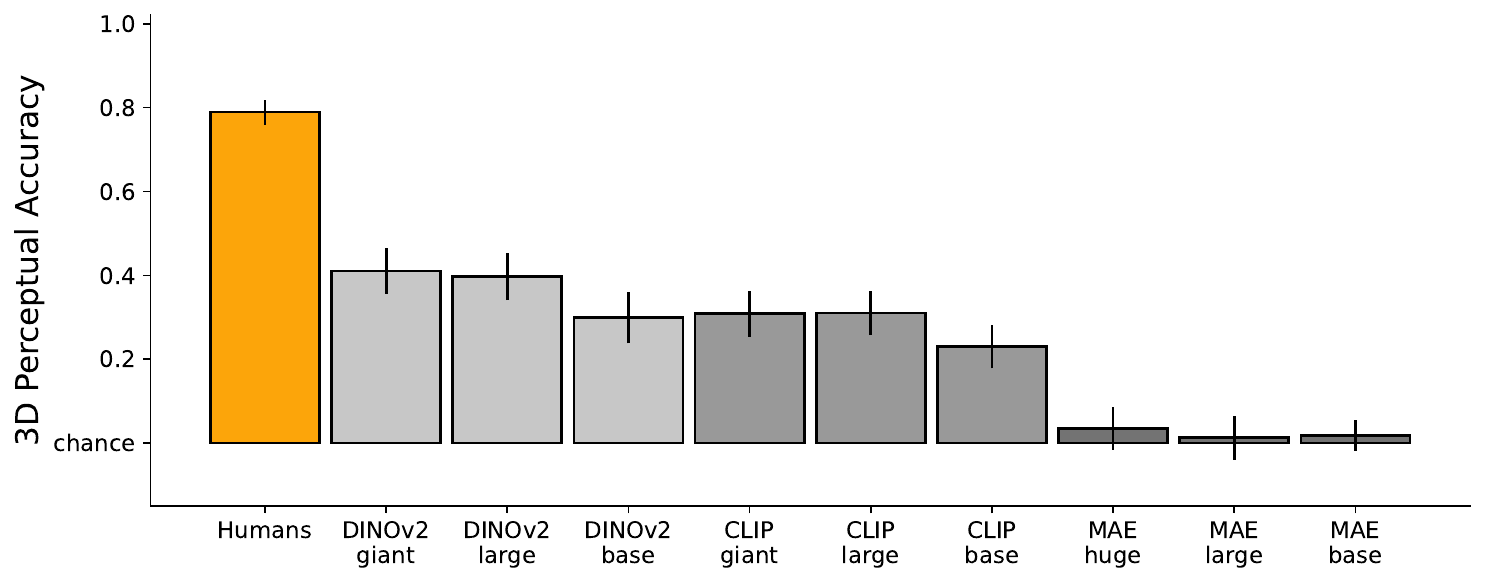}
\end{center}
\vspace{-5mm}
\caption{\textbf{Standard vision models fail on 3D shape inferences.} We evaluate the ability of large vision models on the MOCHI benchmark (\cite{bonnen2024evaluating}) using a cosine distance metric. For each trial, the image with the lowest average cosine distance to all other images is selected as the oddity. We compare to ground truth and average across all trials. Evaluating DINOv2, CLIP, and MAE models at multiple scales, we find that no model approaches human-level performance. Even the best-performing model (DINOv2-giant) performs at half the accuracy of humans, indicating that the geometric structure of these feature spaces does not naturally separate 3D object identity. }
\label{distance_metrics}
\end{figure}

\begin{figure}[ht!]
\begin{center}
\includegraphics[width=.7\textwidth]{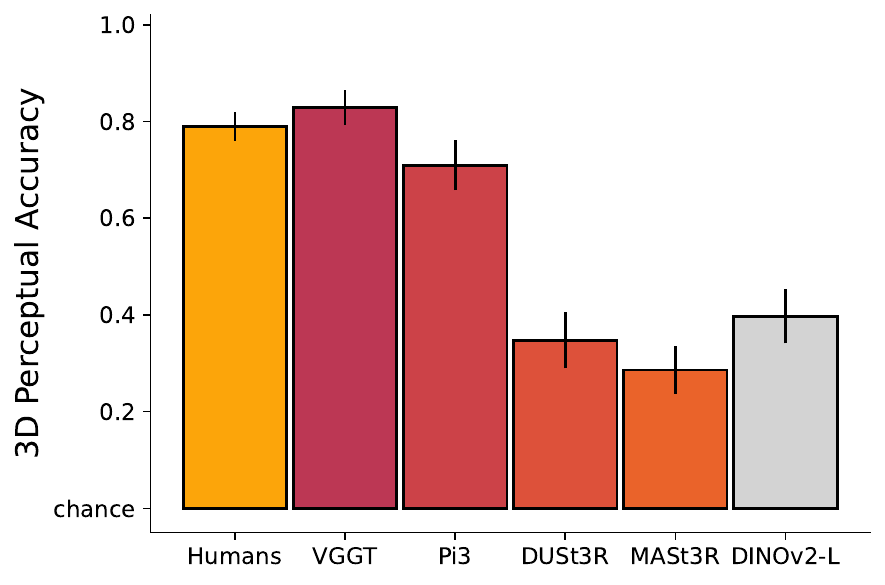}
\end{center}
\vspace{-5mm}
\caption{\textbf{3D shape performance of all vision models evaluated.} We begin our analysis by evaluating the performance of state of the art multi-view transformers on the MOCHI (\cite{bonnen2024evaluating}) benchmark. This includes VGGT (\cite{wang2025vggt}), Pi3 (\cite{wang2025pi}), MAST3R (\cite{leroy2024grounding}), DUST3R (\cite{wang2024dust3r}), as well as a zero-shot evaluation of DINOv2 (\cite{oquab2023dinov2}). Because VGGT is the only model that achieves human-level performance on this 3D shape inference benchmark we restrict all subsequent analyses to this model. }
\label{allmodels}
\end{figure}

\begin{figure}[ht!]
\begin{center}
\includegraphics[width=.7\textwidth]{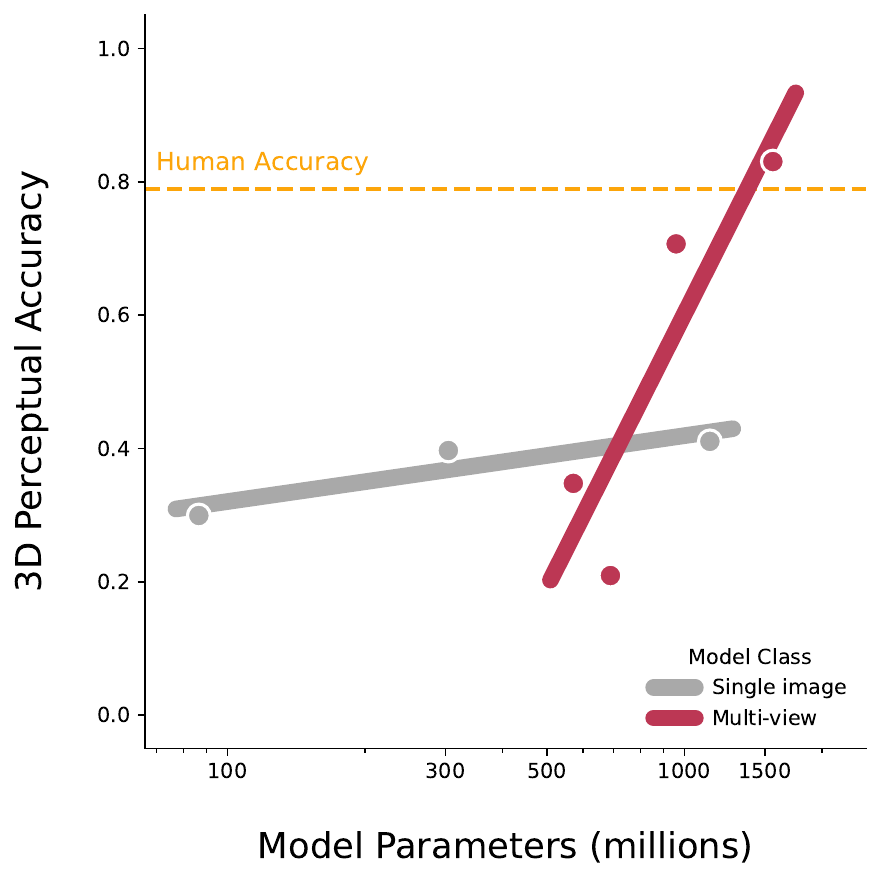}
\end{center}
\vspace{-5mm}
\caption{\textbf{Visualizing model performance as a function of model size.} }
\label{allmodels}
\end{figure}

\begin{figure}[ht!]
\begin{center}
\includegraphics[width=1\textwidth]{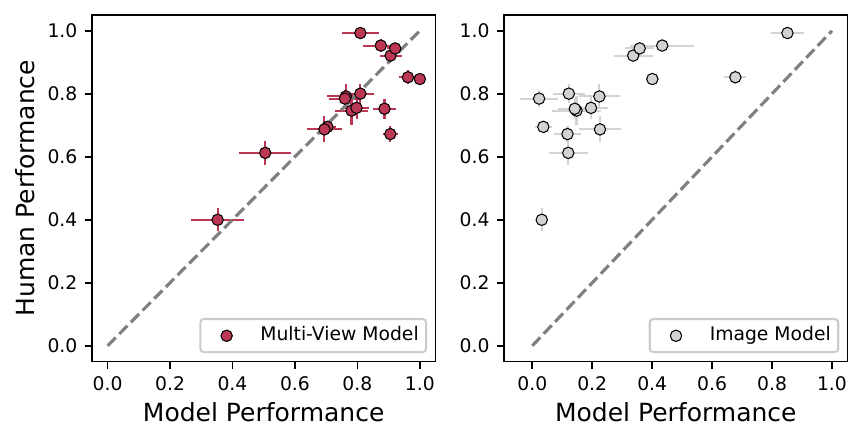}
\end{center}
\vspace{-5mm}
\caption{\textbf{Comparing the relationship between humans and vision models per condition.} Here we visualize the correlation between human and model performance per condition (i.e., averaging across all trials within a given condition) for both VGGT (left) and DINOv2 (right). Human-VGGT performance is largely on diagonal, indicating that VGGT is predicting human-level performance across conditions. In contrast, Human-DINOv2 performance is significantly above the diagonal, indicating that humans are significantly outperforming the DINOv2 across conditions. }
\label{conditioncorrelation}
\end{figure}

\begin{figure}[ht!]
\begin{center}
\includegraphics[width=1\textwidth]{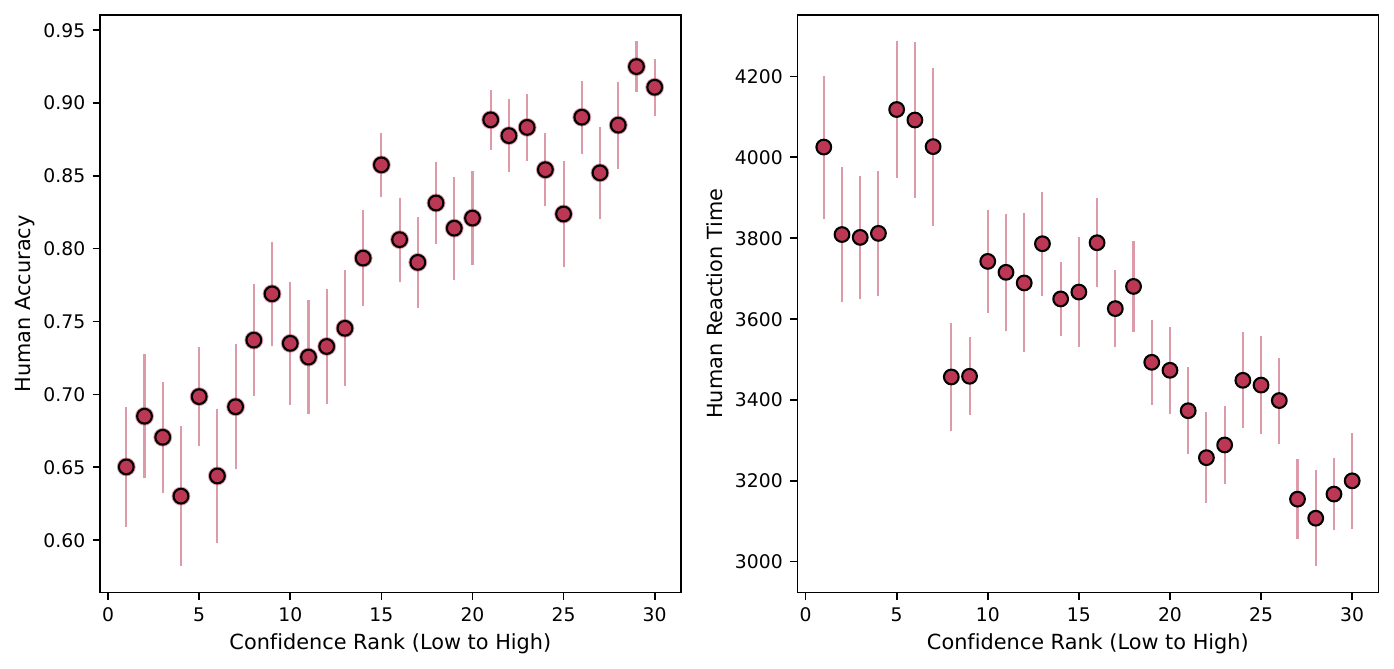}
\end{center}
\vspace{-5mm}
\caption{\textbf{Direct relationship between model confidence and human accuracy/RT.} Because human accuracy and reaction time are correlated, the ability to predict one measure typically implies the ability to predict the other. Here we visualize the relationship between model confidence and both human error patterns (left) and reaction time (right), demonstrating this clear correspondence. It is in part because of this circularity that we developed an independent model readout to evaluate the model's ability to predict reaction time data; model `solution time' is independent of confidence measures.}
\label{readouts}
\end{figure}


\begin{figure}[ht!]
\begin{center}
\includegraphics[width=1\textwidth]{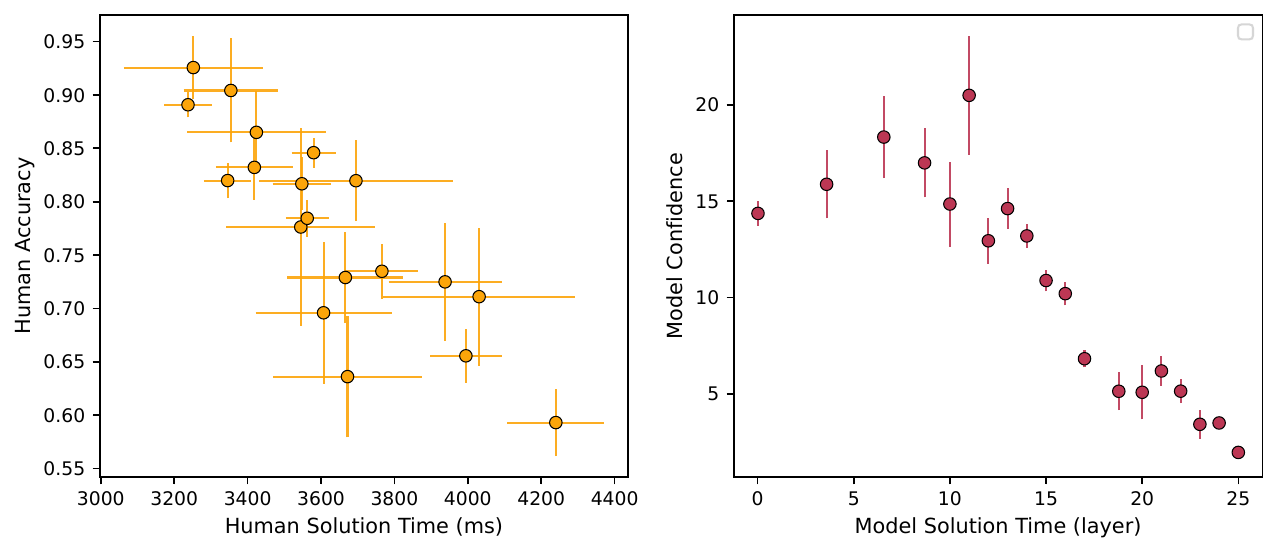}
\end{center}
\vspace{-5mm}
\caption{\textbf{The relationship between accuracy and reaction time, ordered by model solution time.} To independently corroborate the utility of model solution time as a predictor of human reaction time, here we use the binned trials identified by our solution layer metric in order to visualize the relationship between human reaction time (x axis, left) and human accuracy (y axis, left): there is a clear relationship between RT and accuracy when grouping trials according to model solution layer, corroborating the validity of this measure. We also plot the relationship between model solution time (x axis, right) and model confidence (y axis, right).}
\label{readouts}
\end{figure}

\begin{figure}[ht!]
\begin{center}
\includegraphics[width=1\textwidth]{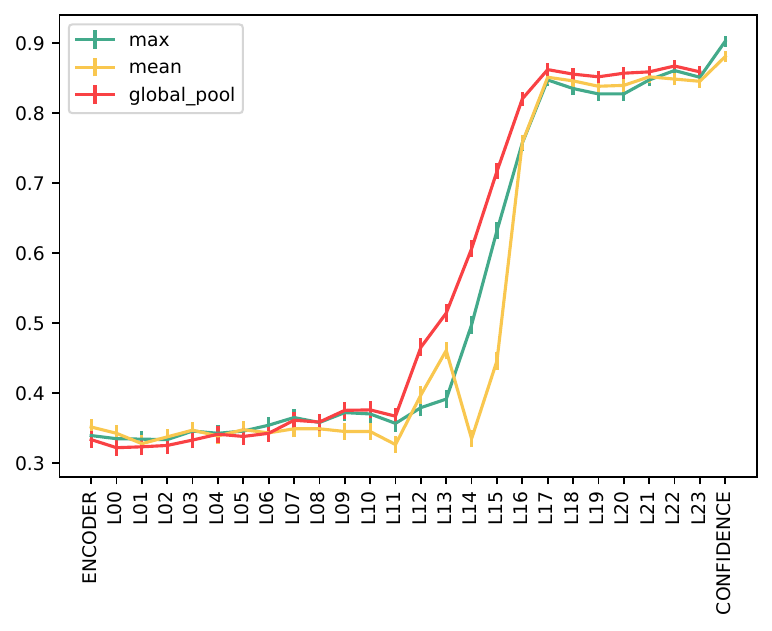}
\end{center}
\vspace{-5mm}
\caption{\textbf{Evaluating the performance of VGGT across multiple layers/readouts}. In addition to the confidence-based performance metric, we also develop layer-wise performance metrics to observe how performance evolves across the model's layer-by-layer processes using the max (green), mean (yellow) and global average pooling of the layer features. For each metric we simple determine, at each layer, whether A and A$'$ are more similar in representational space than A and B.}
\label{readouts}
\end{figure}

\begin{figure}[ht!]
\begin{center}
\includegraphics[width=1\textwidth]{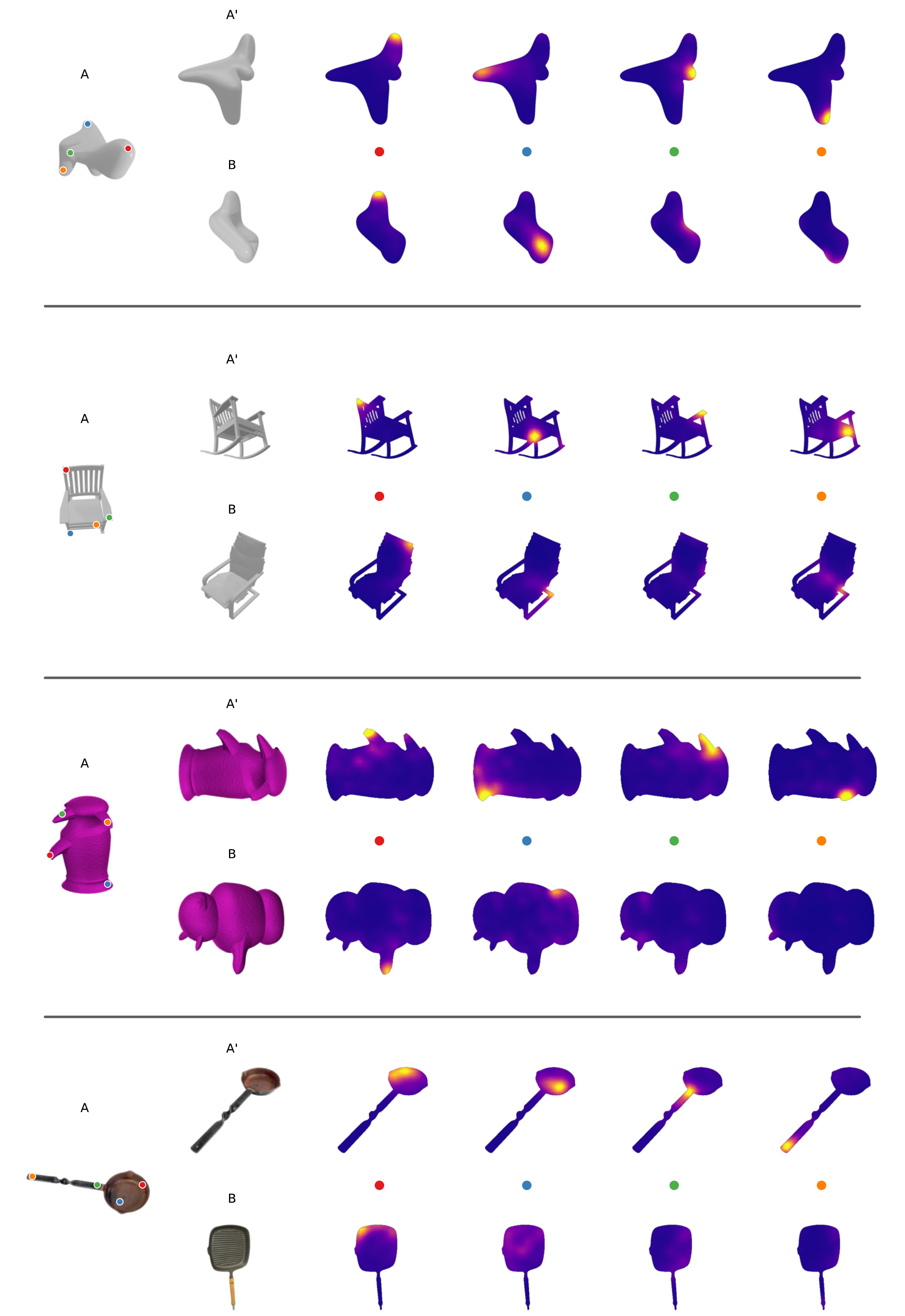}
\end{center}
\vspace{-5mm}
\caption{\textbf{Pattern of global attention across different objects in layer 15.}}
\label{seeattention_middle_four}
\end{figure}

\end{document}